\documentclass{article}

\usepackage{PRIMEarxiv}

\usepackage[utf8]{inputenc} 
\usepackage[T1]{fontenc}    
\usepackage{hyperref}       
\usepackage{url}            
\usepackage{booktabs}       
\usepackage{amsfonts}       
\usepackage{nicefrac}       
\usepackage{microtype}      
\usepackage{lipsum}
\usepackage{graphicx}
\graphicspath{{media/}}     
\usepackage{natbib}
  
\title{Scalable Machine Learning Analysis of Parker Solar Probe Solar Wind Data}

\author{%
 Daniela Martin\\
 University of Delaware \\
 \texttt{dmartinv@udel.edu}
 \And
 Connor O'Brien \\
 Boston University \\
 \texttt{obrienco@bu.edu} \\
 \And
 Valmir P. Moraes Filho \\
 Catholic University of America \\
 \texttt{moraesfilho@cua.edu} \\
 \And
 Jinsu Hong \\
 Georgia State University \\
 \texttt{jhong36@gsu.edu} \\
 \And
 Jasmine R. Kobayashi\\
 Southwest Research Institute \\
 \texttt{jasmine.kobayashi@swri.org} \\
 \And
 Evangelia Samara \\
 NASA Goddard Space Flight Center \\
 \texttt{evangelia.samara@nasa.gov} \\
 \And
 Joseph Gallego \\
 Drexel University \\
 \texttt{jg3959@drexel.edu} \\
}

\begin{document}
\maketitle

\begin{abstract}
We present a scalable machine learning framework for analyzing Parker Solar Probe (PSP) solar wind data using distributed processing and the quantum-inspired Kernel Density Matrices (KDM) method. The PSP dataset (2018--2024) exceeds 150 GB, challenging conventional analysis approaches. Our framework leverages Dask for large-scale statistical computations and KDM to estimate univariate and bivariate distributions of key solar wind parameters, including solar wind speed, proton density, and proton thermal speed, as well as anomaly thresholds for each parameter. We reveal characteristic trends in the inner heliosphere, including increasing solar wind speed with distance from the Sun, decreasing proton density, and the inverse relationship between speed and density. Solar wind structures play a critical role in enhancing and mediating extreme space weather phenomena and can trigger geomagnetic storms; our analyses provide quantitative insights into these processes. This approach offers a tractable, interpretable, and distributed methodology for exploring complex physical datasets and facilitates reproducible analysis of large-scale in situ measurements. Processed data products and analysis tools are made publicly available to advance future studies of solar wind dynamics and space weather forecasting. The code and configuration files used in this study are publicly available to support reproducibility.
\end{abstract}


\section{Introduction}
The Parker Solar Probe (PSP), launched in 2018, aims to unveil the mechanisms driving the heating and acceleration of the solar wind \cite{raouafi2023parker}. This work focuses on analyzing PSP measurements of solar wind plasma properties obtained by the SWEAP instrument \cite{fox_solar_2016}. The dataset spanning 2018 $-$ 2024 exceeds 150 GB, presenting challenges for scalable and interpretable analysis of high-volume time series data. Prior studies seeking to determine the overall properties of the solar wind from data taken in-situ have been based on smaller datasets with poorer temporal resolution in more limited spatial regimes \cite{ma_statistical_2020, wilson_iii_statistical_2018}, and focus on basic statistical techniques such as binning and averaging.

Traditional approaches, including feature extraction, probabilistic models, and deep learning methods, face limitations when applied to solar wind measurements at this scale \cite{henderson2021empirical,jadon2021challenges}. For instance, kernel density estimation (KDE) \cite{parzen1962estimation} becomes computationally prohibitive, while normalizing flows \cite{rezende2015variational,normalizationflows_gallego_2023} and autogressive models are often difficult to train. Variational autoencoders only approximate the underlying distribution, and implicit generative models such as Generative Adversarial networks \cite{generativeadversarialnetworks_goodfellow_2014} or diffusion models \cite{diffusion_croitoru_2023} do not provide an explicit density function. To address these challenges, we develop a distributed framework using Dask \cite{dask} and apply the quantum-inspired Kernel Density Matrices (KDM) method \cite{gonzalez2023kernel} for scalable density estimation and anomaly detection. The computation of the statistical quantities are parallelized, while not a major contribution for Dask itself, represents a practical application for the heliophysics community. 

Our contributions are threefold: (1) statistical characterization of solar wind properties using distributed processing; (2) application of KDM to uncover multi-parameter distributions and relationships in large PSP solar wind datasets; and (3) open-source tools for analyzing large in situ datasets.

\section{Background}
\cite{gonzalez2023kernel} define KDM over a set $\mathbb{X}$ as a triplet $\rho=(\mathbf{C}, \mathbf{p}, \theta)$ where $\rho$ represents a density matrix defined by 
\[
f_\rho(\mathbf{x}) = \sum_{\mathbf{x}^{(i)} \in \mathbf{C}} p_i k_\theta^2(\mathbf{x}, \mathbf{x}^{(i)}),
\]
Here, $\mathbf{C}$ is the set of components, $\mathbf{x}$ are the original points, and $\mathbf{x}^{(i)}$ is the $i$-th component of $\mathbf{C}$ within $\mathbb{X}$. Each component has an associated mixture weight $p_i\in\mathbb{R}$, representing its probability. The kernel function $k_{\theta}: \mathbb{X}\times\mathbb{X}\rightarrow\mathbb{R}$ is defined such that $k(\mathbf{x},\mathbf{x})=1$. Using a Gaussian kernel, each component $\mathbf{x}^{(i)}$ is assumed to follow a Gaussian distribution with a normalization constant. Components define a categorical distribution, where each category corresponds to a normal distribution with mean equal to the component value and standard deviation $\sigma/2$, i.e. $\mathcal{N}(x^{(i)}, \sigma/2)$. 

The parameters $(\mathbf{C}, \mathbf{p}, \theta)$ are learned by maximizing the log-likelihood of the observed data:
\[
\max_{\mathbf{C}, \mathbf{p}, \theta} \sum_{i=1}^{\ell} \log \hat{f}_{\rho}(x_i),
\]
where $\hat{f}_{\rho}$ is the projection function associated with the KDM density matrix $\rho$ of the component $\mathbf{x}_i$. The model is trained using automatic differentiation. This approach allows us to compute both univariate and joint distributions, enabling subsequent statistical analyses and anomaly detection. The model can also be used to sample new points from the learned distribution.

\section{Experimental Setup \label{sec:experiments}}

\textbf{Dataset.} The Parker Solar Probe (PSP) is a mission launched on Aug. 12, 2018, by NASA, which has completed 24 orbits as of Jun. 19, 2025. Its primary goal is to better understand the structure and dynamics of the inner corona of the Sun, where the solar wind is heated and accelerated. Determining the processes that cause the acceleration of energetic particles is of great importance to the heliophysics community \cite{raouafi2023parker}. PSP carries four instruments: FIELDS (measuring electric and magnetic fields), SWEAP (sampling charged particles such as electrons, protons, and alpha particles), WISPR (imaging the solar corona), and IS$\odot$IS (sampling high-energy particles) \cite{fox_solar_2016}. In this work, we focus on SWEAP data to analyze the plasma properties of the solar wind \cite{kasper_solar_2016, whittlesey_solar_2020, case_solar_2020}.

The SWEAP data must be preprocessed before it can be used for this study. The SWEAP data are first downloaded as CDF files and converted to Zarr format \cite{zarr-github2025} to enable distributed computation. Fill values due to data outages are flagged and removed, and PSP's radial distance to the Sun is computed and stored alongside the plasma data. The plasma parameters of interest are derived from the raw instrument data either by fitting several Gaussians to the particle distributions measured by the instruments or by taking moments of the particle distributions (see \cite{kasper_solar_2016, whittlesey_solar_2020, case_solar_2020} for details). Here we use data derived using the fit technique due to their improved temporal coverage.

\textbf{Processing Architecture.} Our processing architecture comprises two parts: (1) distributed processing using Dask and (2) KDM for estimating univariate and joint distributions of PSP solar wind parameters. The PSP data are processed using Dask Array across multiple cores and nodes. The analysis was performed in two stages: first, Dask was used to compute statistical metrics from the PSP solar wind dataset; second, KDM was applied to generate univariate and bivariate distributions of parameters such as solar wind speed, proton density, and proton thermal speed. These parameters were selected because they represent the primary physical quantities governing solar wind behavior and provide complementary information on both its bulk motion and thermal state, making them ideal for studying correlations and variability in the inner heliosphere. All analyses were conducted as a function of heliocentric distance, ranging from 0 to 1 astronomical units (AU) in increments of 0.1 AU.

\textbf{Codebase. } The full codebase supporting this work is available at \url{https://github.com/spaceml-org/PSP-KDM}.

\textbf{Hardware.} All experiments were conducted on Google Cloud Platform using a c2-standard-8 VM instance (8 vCPUs, 32 GB RAM) with 1 TB SSD storage and a Nvidia L4 (24 GB V-GPU).
\begin{figure}[htbp!]
        \includegraphics[width=0.33\textwidth]{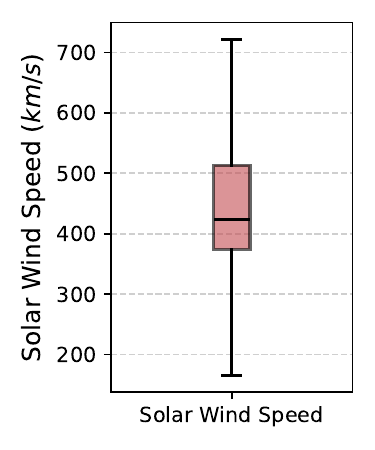}
        \includegraphics[width=0.33\textwidth]{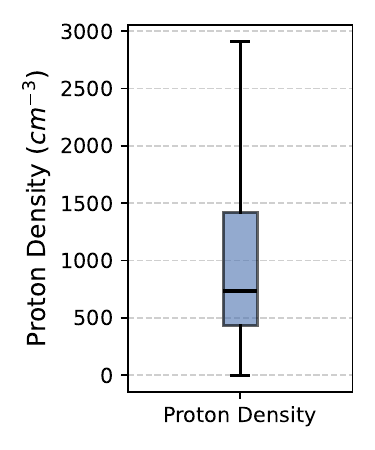}
        \includegraphics[width=0.33\textwidth]{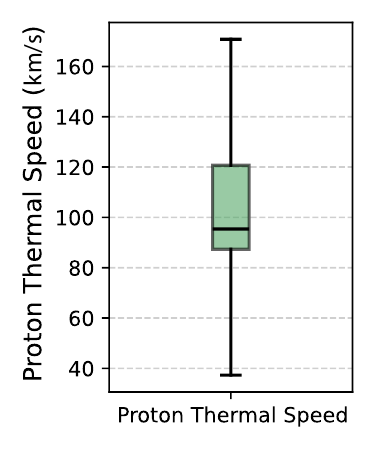}
    \caption{Boxplots of three PSP solar wind parameters from 2018 to 2024: (left) solar wind speed [km/s], (middle) proton density [cm\textsuperscript{-3}], and (right) proton thermal speed [eV]. 
    }
    \label{fig:boxplot_params}
\end{figure}
\section{Results and Discussion}

Several statistics computed with Dask reveal notable features in the PSP solar wind data. Figure~\ref{fig:boxplot_params} shows boxplots for solar wind speed, proton density, and proton thermal speed for the entire dataset (2018 to 2024). Extreme solar wind speed values range from 100 km/s to 1980 km/s, likely caused by instrument artifacts, with a median of 423 km/s. Proton density spans 0.00657$-$2,560 cm\textsuperscript{-3}, from a possible ``disappearing solar wind'' event \cite{fowler_disappearing_2024} at the lower bound to either a genuine coronal feature or an outlier at the upper bound. Its median value of 734 cm\textsuperscript{-3}, agrees with previous PSP observations \cite{maruca2023trans}. The proton thermal speed (\textit{wp\_fit}) ranges from 4 km/s (0.084 $eV$, a potential outlier, to 109,900 km/s (63 $MeV$), values that almost certainly reflect instrument error. Its median of 95.4 km/s (46 $eV$) aligns with typical coronal and inner heliosphere conditions \cite{ma_statistical_2020}.

Table~\ref{tab:hyperparams} summarizes the KDM hyperparameters chosen for all experiments, including the number of components, Gaussian kernel width ($\sigma$), and learning rate.

\begin{table}[htbp!]
\centering
\caption{KDM hyperparameters. The number of components, selected empirically, provides a balance between model flexibility and physical interpretability. Fewer components tend to underfit the data, yielding smoother density estimates with slightly higher mean error but improved generalization, whereas an excessive number produces spikier densities that tend to overfit the data.}
\begin{tabular}{|c|c|}
\hline
\textbf{Hyperparameter} & \textbf{Value} \\
\hline
Number of Components & 400, 800, 1600 \\
\hline
$\sigma$ & 0.1 (trainable) \\
\hline
Learning Rate & $10^{-3}$ \\
\hline
\end{tabular}
\label{tab:hyperparams}
\end{table}

\begin{figure}[htbp]
        \includegraphics[width=0.45\textwidth]{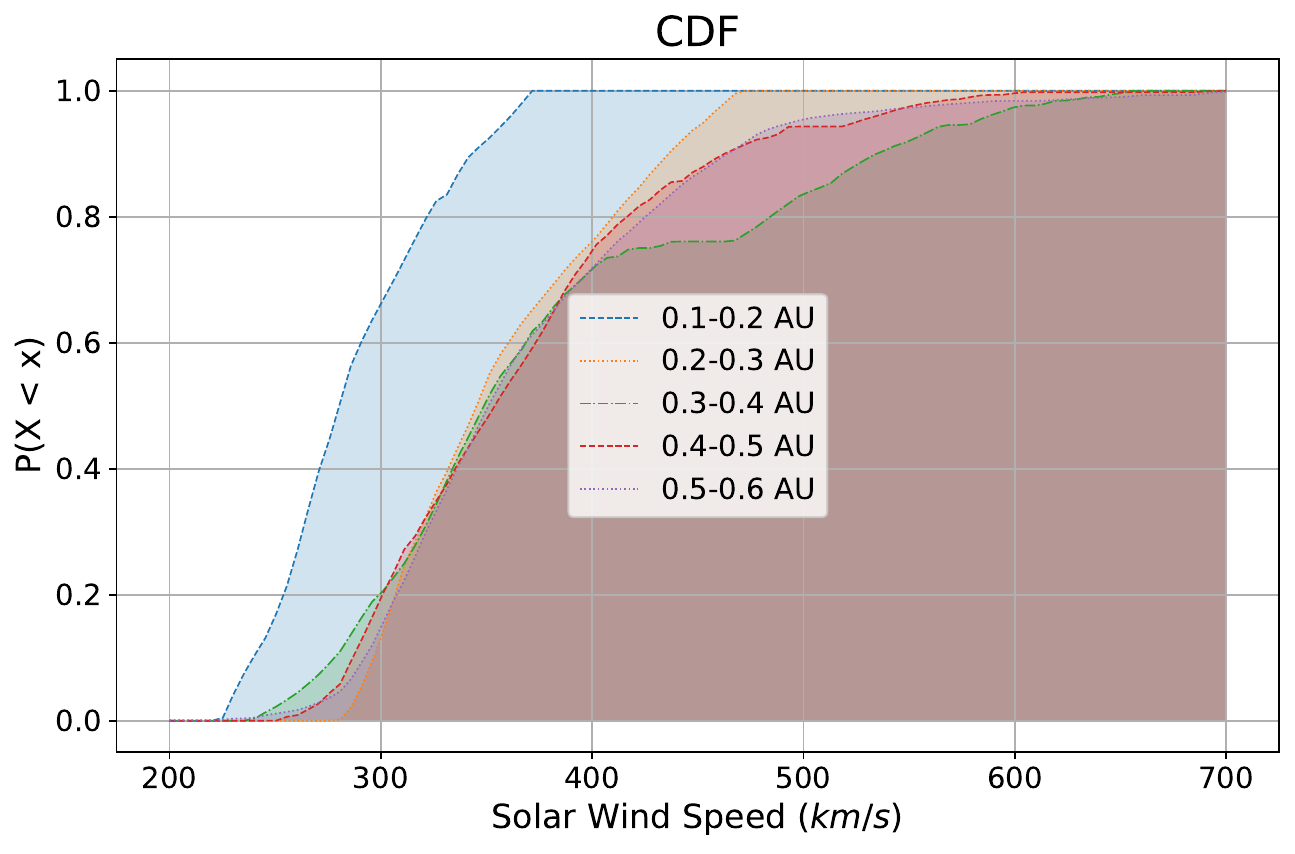}
        \includegraphics[width=0.46\textwidth]{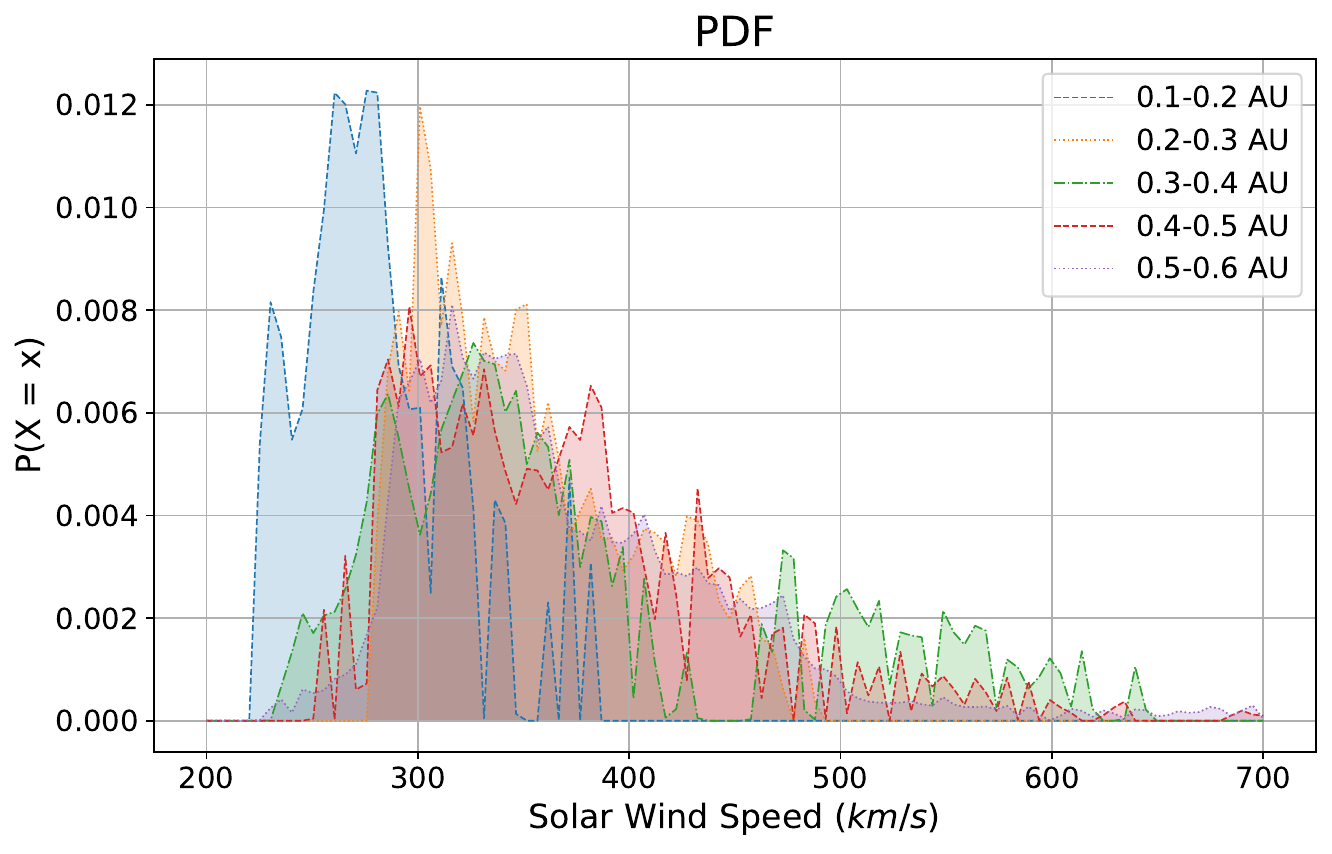}
    \caption{Univariate distributions of PSP's solar wind speed. \textbf{Left:} Cumulative Distribution Function (CDF). \textbf{Right:} Probability Density Function (PDF). Each curve corresponds to a different heliocentric distance (0-0.6 AU), as indicated in the legend. Distances beyond 0.6 AU are omitted due to sparse measurements and increased noise.}
    \label{fig:univ-dist-cdf-pdf-speed}
\end{figure}

Figure \ref{fig:univ-dist-cdf-pdf-speed} shows the univariate distributions of PSP solar wind speed from 0.1 to 0.6 AU in increments of 0.1 AU. Although the analyses covered the full range of 0 to 1 AU, only the subset up to 0.6 AU is shown, since beyond this point the distributions become sparse and noisy due to fewer PSP measurements, which produced spiky curves that are not statistically robust. Our focus on 0.1 to 0.6 AU highlights the near-Sun region where data density is highest and the physical insights are therefore more reliable. The cumulative distribution function (CDF) indicates that solar wind speeds are lower closer to the Sun, consistent with coronal measurements before acceleration \cite{patel_direct_2025}. Between 0.2 (orange) and 0.4 AU (red), speeds appear relatively uniform, though at 0.3 AU (green) both fast and slow solar streams emerge.

The probability density function (PDF) in Figure~\ref{fig:univ-dist-cdf-pdf-speed} exhibits a right-skewed distribution at all distances. At 0.2 AU (orange), the distribution is leptokurtic, reflecting relatively homogeneous velocities, whereas at 0.3 AU (green) it becomes platykurtic, indicating broader variations. The presence of a large high-speed tail at 0.3 AU, particularly above 450 km/s, suggests increased variability in the solar wind speed, likely associated with transient events or dynamic coronal structures. Together, these trends provide insights into the acceleration and variability of the solar wind in the inner heliosphere.

\begin{figure}[htbp]
        \includegraphics[width=0.5\textwidth]{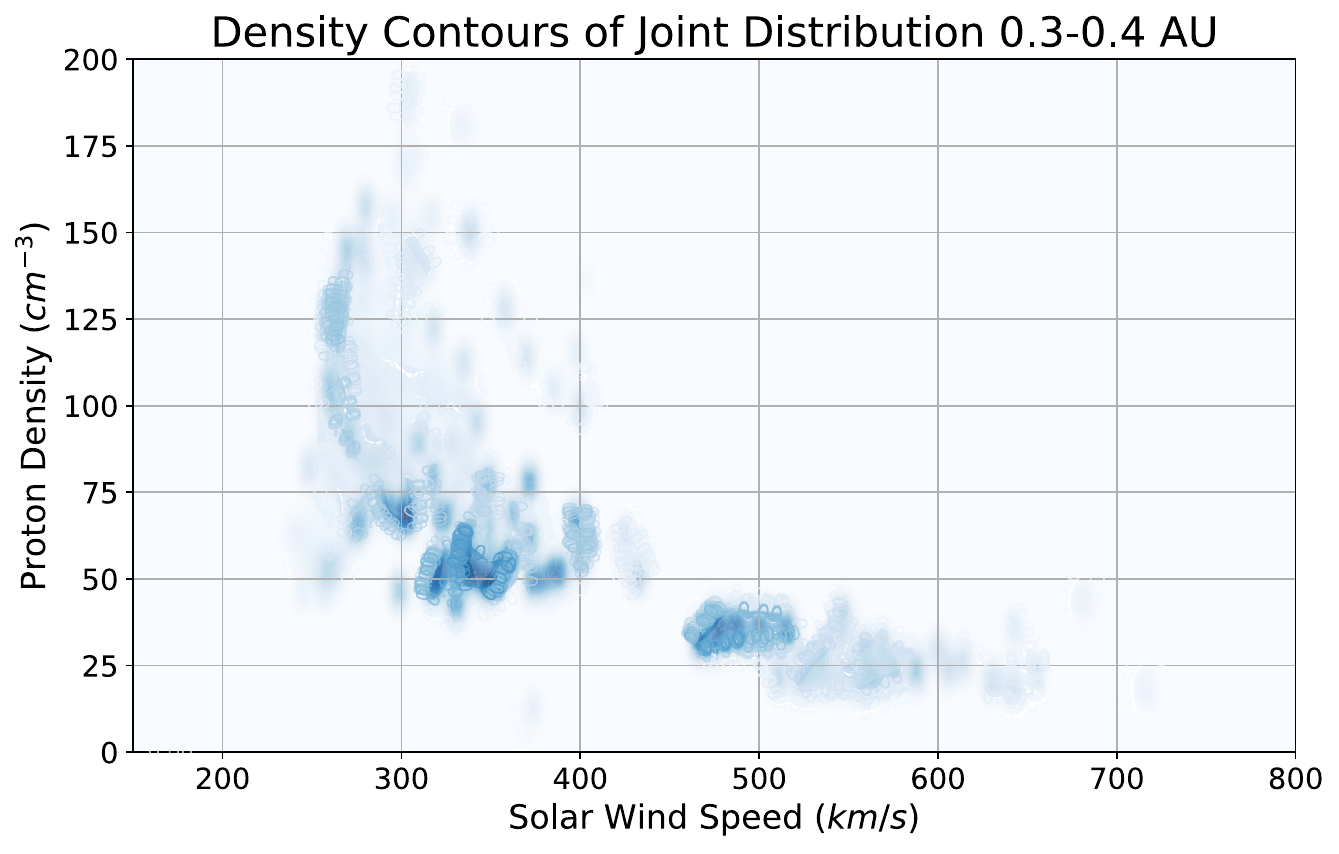}
        \includegraphics[width=0.5\textwidth]{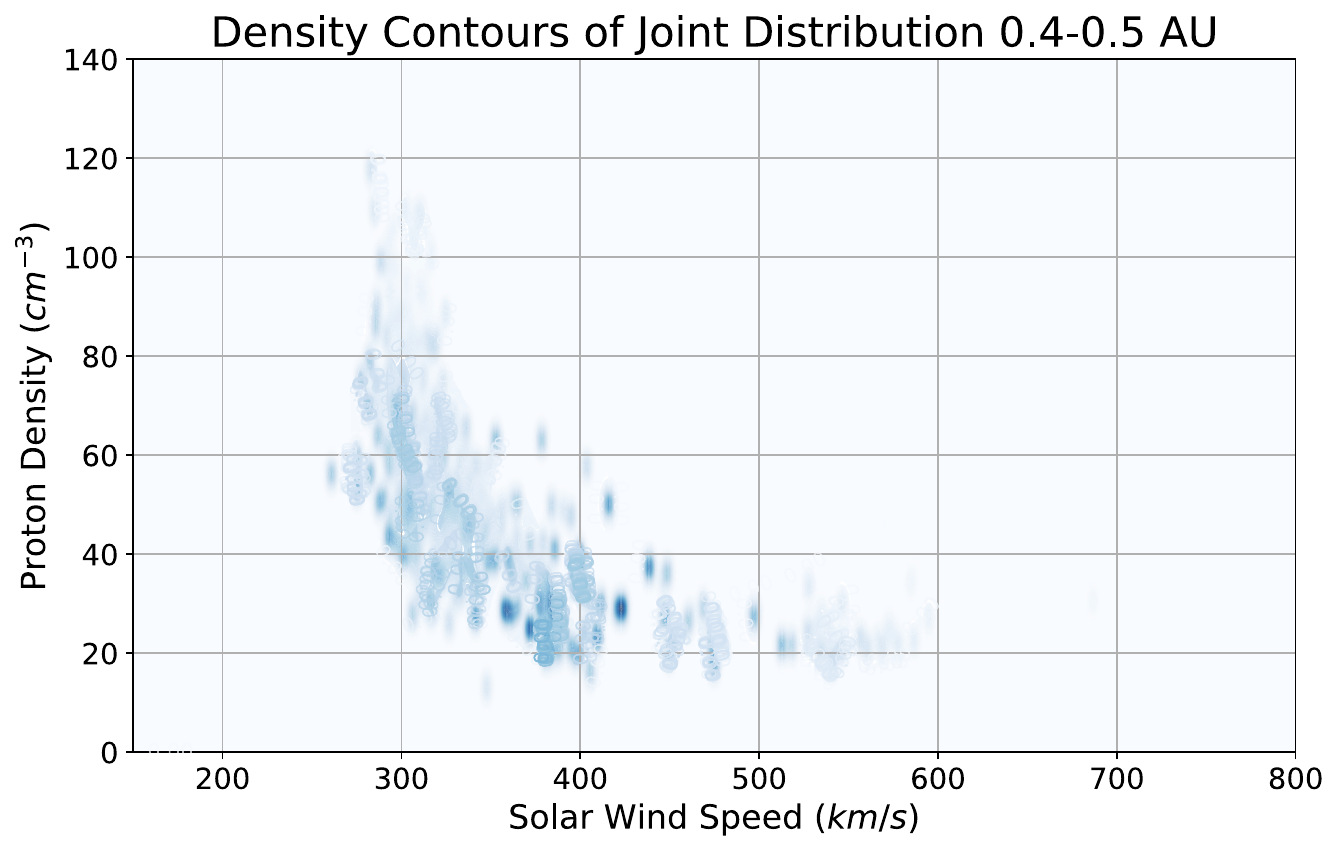}
    \caption{Bivariate distributions of PSP's solar wind speed versus proton density at different heliocentric distances (AU). \textbf{Left panel:} 0.3-0.4 AU. \textbf{Right panel:} 0.4-0.5 AU.}
    \label{fig:bivariate-dist-au}
\end{figure}

Figure \ref{fig:bivariate-dist-au} illustrates the bivariate distribution of solar wind speed versus proton density at different heliocentric distances. The left panel shows the distribution between 0.3 and 0.4 AU, revealing a clear hyperbolic inverse relationship consistent with a ``textbook'' understanding of solar wind structures \cite{kallenrode_space_2010}: as solar wind speed increases, proton density decreases. Notably, there is considerable variability in the proton density when the solar wind speed is between 250 and 350 km/s, which may indicate multiple sources or types of solar wind within this slower population. The right panel presents the same relationship for 0.4 to 0.5 AU, where the hyperbolic inverse trend remains. The primary difference is a reduction in the maximum observed solar wind speed, which decreases from approximately 700 km/s at 0.3$-$0.4 AU range to around 600 km/s at 0.4$-$0.5 AU, along with a decrease in the maximum proton density from 200 cm\textsuperscript{-3} to 120 cm\textsuperscript{-3}. The density reduction is consistent with rarefaction of the solar wind as it expands radially \cite{russel_2016, timar20243d}.

We focus on the univariate distributions of solar wind speed and the bivariate distributions of solar wind speed versus proton density in the main text because these parameters are the most representative of solar wind dynamics. The solar wind speed captures the bulk flow behavior, while its relationship with proton density illustrates key physical phenomena such as the inverse correlation observed in the inner heliosphere. Other parameters and combinations are reported in the supplemental material. The supplemental material provides the corresponding univariate and bivariate distributions for all other solar wind parameters and combinations.

\section{Conclusion}
We present a scalable and interpretable framework for analyzing Parker Solar Probe solar wind measurements, combining distributed computation with the quantum-inspired KDM method. By computing univariate and bivariate distributions across heliocentric distances, our approach captures key physical relationships, including the inverse correlation between solar wind speed and proton density, as well as variability in kurtosis and skewness reflecting plasma dynamics in the inner heliosphere. Limiting the main analysis to solar wind speed and proton density allows us to focus on the most physically relevant quantities, while additional distributions and parameter combinations are provided in the supplemental material.

Our methodology is efficient, adaptable to other high-volume space physics datasets, and facilitates reproducible analysis of large-scale in situ measurements. The framework also enables the identification of anomalies and extreme events, highlighting its potential for uncovering physically meaningful patterns and supporting future space weather studies.

\section*{Broader Impact}

Understanding solar wind phenomena is of paramount importance for deciphering the fundamental drivers of extreme solar events such as coronal mass ejections. Moreover, comparing observations from the Parker Solar Probe (PSP) with established physical models enhances confidence in forecasting frameworks developed using PSP data.

\section*{Acknowledgements}
This work is a research product of Heliolab (heliolab.ai), an initiative of the Frontier Development Lab (FDL.ai). FDL is a public–private partnership between NASA, Trillium Technologies (trillium.tech), and commercial AI partners including Google Cloud and NVIDIA. Heliolab was designed, delivered, and managed by Trillium Technologies Inc., a research and development company focused on intelligent systems and collaborative communities for Heliophysics, planetary stewardship and space exploration. We gratefully acknowledge Google Cloud for extensive computational resources enabled through VMware. This material is based upon work supported by NASA under award No. 80GSFC23CA040. Any opinions, findings, and conclusions or recommendations expressed are those of the author(s) and do not necessarily reflect the views of the National Aeronautics and Space Administration.

\bibliographystyle{unsrt}  
\bibliography{references}

\end{document}